\pdfoutput=1

\documentclass{bmvc2k}
\usepackage{graphicx}
\usepackage{amssymb}
\usepackage{dblfloatfix}    
\usepackage{float}
\usepackage{outlines}
\usepackage{titlesec}
\graphicspath{ {./figures/} }

\title{On Modality Bias in the TVQA Dataset}

\addauthor{Thomas Winterbottom}{thomas.i.winterbottom@durham.ac.uk}{1}
\addauthor{Sarah Xiao}{hong.xiao@durham.ac.uk}{2}
\addauthor{Alistair McLean}{alistair@carbonrmp.com}{3}
\addauthor{Noura Al Moubayed}{noura.al-moubayed@durham.ac.uk}{1}

\addinstitution{
 Department of Computer Science\\
 Durham University\\
 Durham, UK
}
\addinstitution{
 Business School\\
 Durham University\\
 Durham, UK
}
\addinstitution{
 Carbon (AI) \\
 Middlesbrough, UK
}
\runninghead{Winterbottom, Xiao, McLean, Al Moubayed}{On Modality Bias in TVQA}

\titleformat*{\section}{\normalfont\Large\bfseries\flushleft\color{bmv@sectioncolor}}
\titleformat*{\subsection}{\normalfont\large\bfseries\flushleft\color{bmv@sectioncolor}}
\begin{document}

\maketitle
\begin{abstract}
TVQA is a large scale video question answering (video-QA) dataset based on popular TV shows. The questions were specifically designed to require ``both vision and language understanding to answer''. In this work, we demonstrate an inherent bias in the dataset towards the textual subtitle modality. We infer said bias both directly and indirectly, notably finding that models trained with subtitles learn, on-average, to suppress video feature contribution. Our results demonstrate that models trained on only the visual information can answer $\sim$45\% of the questions, while using only the subtitles achieves $\sim$68\%. We find that a bilinear pooling based joint representation of modalities damages model performance by 9\% implying a reliance on modality specific information. We also show that TVQA fails to benefit from the RUBi modality bias reduction technique popularised in VQA. By simply improving text processing using BERT embeddings with the simple model first proposed for TVQA, we achieve state-of-the-art results (72.13\%) compared to the highly complex STAGE model (70.50\%). We recommend a multimodal evaluation framework that can highlight biases in models and isolate visual and textual reliant subsets of data. Using this framework we propose subsets of TVQA that respond exclusively to either or both modalities in order to facilitate multimodal modelling as TVQA originally intended.
\end{abstract}

\section{Introduction}
The increased complexity and capability of models on a single given modality (e.g. image, text) has sparked interest in modelling multimodal data, especially over the last five years \cite{DBLP:journals/corr/MalinowskiF14, DBLP:journals/corr/AntolALMBZP15, DBLP:journals/corr/FukuiPYRDR16, AlAmri2019AudioVS, Das2019VisualD, zhang2019multimodal, 8715409}. The visual question answering (VQA) \cite{DBLP:journals/corr/AntolALMBZP15, Gupta2017SurveyOV} task in particular has inspired a wealth of multimodal fusion techniques and models \cite{zhang2019multimodal}, and taxonomic analysis of the latent representations they produce \cite{8715409}. However, videos promise more raw visual content than still images used in VQA, and include temporal dependencies that models could exploit. Multiple video question answering (video-QA) datasets have been developed (MovieQA \cite{MovieQA}, MovieFIB \cite{maharaj2017dataset}, PororoQA \cite{pororoqa}, TGIF-QA \cite{jang-CVPR-2017}, YouTube2Text-QA \cite{DBLP:journals/corr/YeZLCXZ17}, EgoVQA \cite{Fan_2019_ICCV} and TVQA \cite{lei2018tvqa}). The TVQA\footnote{http://tvqa.cs.unc.edu} dataset was designed to address shortcomings in previous datasets. It is relatively large, uses longer clips and realistic video content, and provides timestamps allowing the identification of the subtitles and video frames relevant to a given question. Most notably, the questions were specifically designed to encourage multimodal reasoning, i.e. models built to answer the questions should require both visual \textit{and} textual cues simultaneously. To achieve this, Amazon Mechanical Turk (AMT) workers were asked to design two-part compositional questions with a `main' part (``What was House saying..'') and a `grounding' part (``..before he leaned over the bed?''). The authors claimed this will naturally produce questions that require both visual and language information to answer since ``people often naturally use visual signals to ground questions in time''. Despite these specific efforts to ensure that ``questions require both vision understanding \textit{and} language understanding'', we show that in practice this is not the case. We demonstrate the subtitles are informative enough to answer the majority of questions in TVQA without requiring complementary visual information as intended. We show that 68\% of the questions can be correctly answered using only the subtitles. Adding visual information merely increases the accuracy to 72\%, without subtitles this drops by 27\%. TVQA authors stress the importance of subtitles in video-QA ``because it reflects the real world, where people interact through language''. Though this is true and subtitles significantly improves performance on TVQA, we find their inclusion actively discourages multimodal reasoning and that the subtitles dominate rather than complement the video features. The TVQA+ dataset \cite{lei2019tvqa} is not considered in this study. Despite TVQA+ providing improved timestamp annotations, it is a significantly smaller subset of TVQA. Furthermore, the `visual concept words' collected for TVQA+ sample from the questions and the correct answers. This means that models trained on TVQA+ will be trained to use additional textual hints disproportionately from correct answers \footnote{TVQA+: http://tvqa.cs.unc.edu/download\_tvqa\_plus.html}. This defeats the purpose of video-QA models as it assumes the correct answer is known to the model during feature extraction. The main contributions of this paper are: I) An evaluation framework for multimodal datasets. II) Extensive analysis of the performance of the TVQA model and dataset per modality and feature type, including the relative contributions of each feature, notably finding that models trained with subtitles learn to suppress video feature contribution. III) Demonstrate an inherent reliance in the questions on the subtitles rather than multimodal interactions. IV) State of the art results achieved using the baseline TVQA model by simply boosting its textual reasoning with contextual embeddings using BERT \cite{devlin2018bert}. V) Define data subsets that respond exclusively to a single modality or a combination of modalities. VI) Demonstrate that the model-agnostic RUBi learning strategy (reducing unimodal bias) \cite{Cadne2019RUBiRU} fails to improve TVQA performance, inline with other textually biased datasets. VII) Apply bilinear pooling fusion to the TVQA baseline. The evaluation framework and proposed subsets are available on GitHub\footnote{Available at https://github.com/Jumperkables/tvqa\_modality\_bias}.

\section{Related Work} QA datasets with visual inputs have been key areas for developing multimodal reasoning in deep learning. However, merely collecting a real world dataset inevitably introduces biases into it \cite{Chao2018BeingNB, 10.1145/2509558.2509563, 5995347, tommasi2017deeper}.

\noindent \textbf{Video Question Answering:} One of the earliest practical video-QA datasets is MovieQA \cite{MovieQA}, which provided long video clips but had a relatively low number of clips and QA pairs. MovieFIB \cite{maharaj2017dataset} is a fill-in-the-blank QA dataset. Although the task is simpler than free-language QA, MovieFIB has substantially more QA pairs than MovieQA. PororoQA \cite{pororoqa} is smaller with cartoon-based video clips to simplify the stories. TGIF-QA \cite{jang-CVPR-2017} uses short social media GIFs and emphasises actions and verbs. EgoVQA \cite{Fan_2019_ICCV} uses first person video to emulate the natural behaviour of intelligent agents. The AVSD dataset \cite{AlAmri2019AudioVS} is a visual dialog dataset i.e. multiple rounds of ongoing QA `dialog' about a given video. AVSD offers audio signals and `dialog history'.

\noindent \textbf{VQA Language Bias:} Though VQA benchmarks are specifically designed to be multimodal, they suffer from strong language priors overriding visual information \cite{Agrawal2015VQAVQ}. VQA v2 \cite{goyal2017making} was designed to weaken those priors \cite{Agrawal2015VQAVQ}. For each question, a new image-question-answer triplet is created with the same question, a similar image, and a different answer. These `balancing' questions help mitigate question specific priors. A `changing priors' extension for both versions \cite{Agrawal2018DontJA} was introduced as a harder `diagnostic' version. Without congruent language priors from the training splits to exploit, many VQA models significantly degraded in performance on the test splits \cite{Andreas2015DeepCQ, Agrawal2015VQAVQ, DBLP:journals/corr/FukuiPYRDR16, Lu2015, Yang2015StackedAN}. 

\noindent \textbf{Reducing Bias:} \citet{Agrawal2018DontJA} also introduced the Grounded VQA model (GVQA) that `disentangles' image-feature-recognition and the process of indentifying the space of plausible answers into 2 separate steps. GVQA outperforms previous benchmarks, but the complex design isn't easily transferred to other models. Flexible model-agnostic learning strategies have recently been developed that are trained end-to-end \cite{Ramakrishnan2018OvercomingLP, Cadne2019RUBiRU, Chen2020CounterfactualSS}. \citet{Ramakrishnan2018OvercomingLP} proposed an adversarial difference-in-entropy regularisation scheme that uses a \textit{deliberately} biased question-only model alongside the full model to adversarially reduce language bias learned during training. The RUBi (Reducing Unimodal Bias) learning strategy \cite{Cadne2019RUBiRU} uses a question-only model to apply a learnable mask on the outputs of the full model during training. The mask is intended to dynamically alter the loss by modifying the predictions of the full model. This causes highly biased training examples to make more biased predictions, effectively decreasing (increasing) the loss, and therefore importance, of highly-biased (visually dependent and difficult) training samples.  The CLEVRER dataset \cite{Yi*2020CLEVRER:} aims to both reduce language bias by sampling evenly across question types and mitigate simpler pattern-recognition behaviour. \citet{girdhar2020cater} create the CATER dataset using a homogeneous scene and object set with multiple variable dataset parameters to `fully control' object and scene bias.

\noindent \textbf {TVQA:} The TVQA dataset \cite{lei2018tvqa} is designed to address the shortcomings of previous datasets. It has significantly longer clip lengths than other datasets and is based on TV shows instead of cartoons, giving it realistic video content with simple coherent narratives. It contains over 150k QA pairs. Each question is labelled with timestamps for the relevant video frames and subtitles. The questions were gathered using AMT workers. Most notably, the questions were specifically designed to encourage multimodal reasoning by asking the workers to design two-part compositional questions. The first part asks a question about a `moment' and the second part localises the relevant moment in the video clip i.e. [What/How/Where/Why/
Who/...] --- [when/before/after] ---, e.g. \textit{[What] was House saying [before] he leaned over the bed?}. The authors argue this facilitates questions that require both visual and language information since ``people often naturally use visual signals to ground questions in time''. The authors identify certain biases in the dataset. They find that the average length of correct answers are longer than incorrect answers. They analyse the performance of their proposed baseline model with different combinations of visual and textual features on different question types they have identified. However, they didn't note the substantial performance of their baseline model on either visual or textual features alone. 

\section{Experimental Framework}
The evaluation framework we present here is built on the original TVQA model and is designed to assess the contributions of visual and textual information in a mutlimodal dataset. The goal is to identify any inherent biases towards either modality. As such, we focus the analysis of the model on the processing streams of the visual and textual features, and the use of `context matching'. This provides a powerful tool to assess in isolation the contribution of the individual feature types and any combination of them.

\begin{figure}
    \centering
    \includegraphics[scale=0.1650]{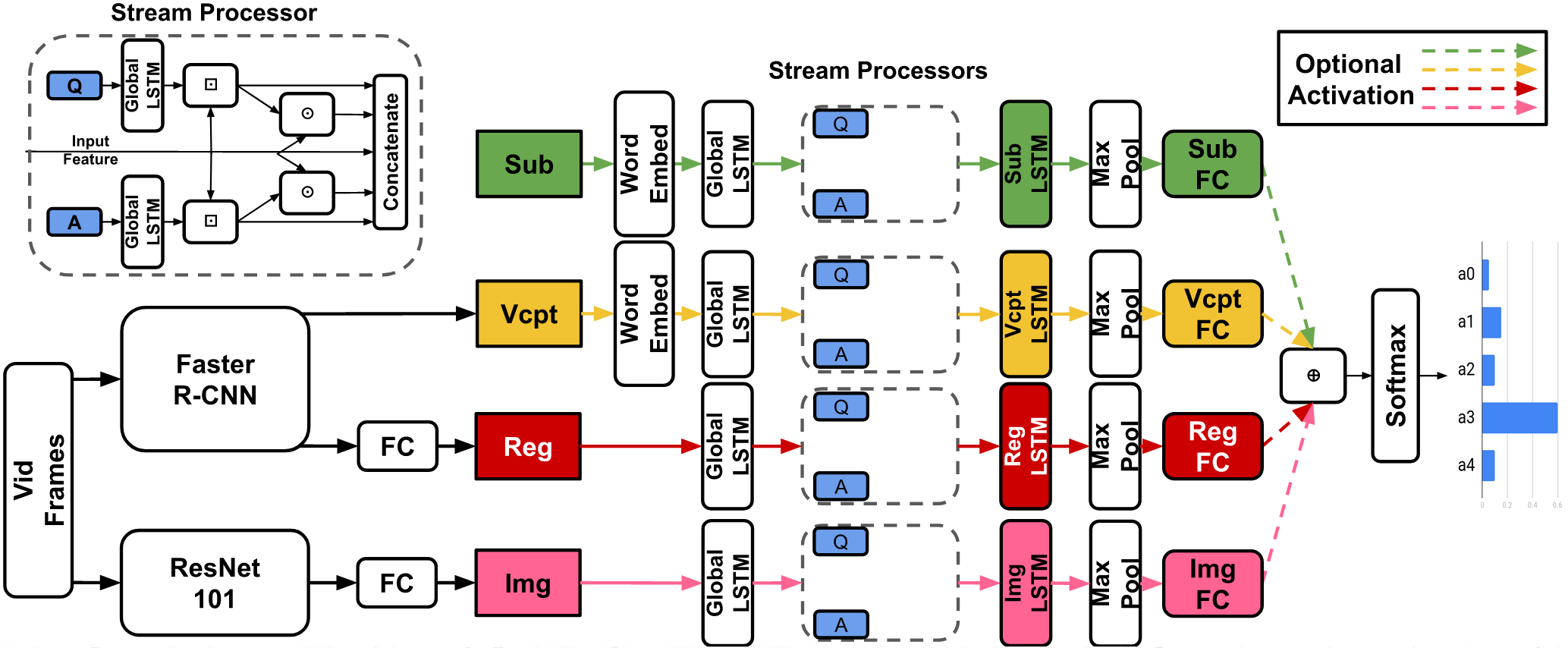}
    \caption{TVQA Model. $\odot$/$\oplus$ = Element-wise multiplication/addition, $\boxdot$ = context matching. Any feature streams may be enabled/disabled.}
    \label{model}
\end{figure}

\noindent \textbf{Model Definition}: The model takes as inputs, I) A question \textit{q} (13.5 words on average), II) Five potential answers $\{\textit{a}_{i}\}_{i=0}^{4}$ (each between 7-23 words), III) A subtitle \textit{S} and video-clip \textit{V} ($\sim$60-90s at 3FPS), and outputs the predicted answer. As the model can either use the entire video-clip and subtitle or only the parts specified in the timestamp, we refer to the sections of video and subtitle used as segments from now on. Figure \ref{model} demonstrates the textual and visual streams and their associated features in the model architecture.

\noindent \textbf{ImageNet Features:} Each frame is processed by a ResNet101  \cite{DBLP:journals/corr/HeZRS15} pretrained on ImageNet \cite{5206848} to produce a 2048-d vector. These vectors are then L2-normalised and stacked in frame order: $V^{img}\in\mathbb{R}^{f\times2048}$ where \textit{f} is the number of frames used in the video segment.

\noindent \textbf{Regional Features:} Each frame is processed by a Faster R-CNN \cite{DBLP:journals/corr/RenHG015} trained on Visual Genome \cite{krishnavisualgenome} in order to detect objects. Each detected object in the frame is given a bounding box, and has an affiliated 2048-d feature extracted. Since there are multiple objects detected per frame (we cap it at 20 per frame), it is difficult to efficiently represent this in time sequences \cite{lei2018tvqa}. The model uses the top-K regions for all detected labels in the segment as in \citet{Anderson2018BottomUpAT} and \citet{Karpathy2015DeepVA}. Hence the regional features are $V^{reg}\in\mathbb{R}^{n_{reg}\times2048}$ where $n_{reg}$ is the number of regional features used in the segment. 

\noindent \textbf{Visual Concepts:} The classes or labels of the detected regional features are called `Visual Concepts'. \citet{inproceedings223} found that simply using detected labels instead of image features gives comparable performance on image captioning tasks. Importantly they argued that combining CNN features with detected labels outperforms either approach alone. Visual concepts are represented as either GloVe \cite{pennington2014glove} or BERT \cite{devlin2018bert} embeddings $V^{vcpt}\in\mathbb{R}^{n_{vcpt}\times 300}$ or $\mathbb{R}^{n_{vcpt}\times 768}$ respectively, where $n_{vcpt}$ is the number of visual concepts used in the segment.

\noindent \textbf{Text Features:} In the evaluation framework, the model encodes the questions, answers, and subtitles using either GloVe  ($\in\mathbb{R}^{300}$) or BERT embeddings ($\in\mathbb{R}^{768}$). Formally,  $q\in\mathbb{R}^{n_q\times d},  \{\textit{a}_{i}\}_{i=0}^{4}\in\mathbb{R}^{n_{a_i}\times d}, S\in\mathbb{R}^{n_s\times d}$ where $n_q, n_{a_i}, n_s$ is the number of words in $q , a_i, S$ respectively and $d=300, 768$ for GloVe or BERT embeddings respectively.

\noindent \textbf{Context Matching:} Context matching refers to context-query attention layers recently adopt-
ed in machine comprehension \cite{DBLP:journals/corr/SeoKFH16, DBLP:journals/corr/abs-1804-09541}. Given a context-query pair, context matching layers return `context aware queries'.

\noindent \textbf{Model and Framework Details:} 
In our evaluation framework, any combination of subtitles or visual features can be used. All features are mapped into word vector space through a tanh non-linear layer. They are then processed by a shared bi-directional LSTM \cite{Hochreiter:1997:LSM:1246443.1246450, graves2005framewise} (`Global LSTM' in Figure \ref{model}) of output dimension 300. Features are context-matched with the question and answers. The original context vector is then concatenated with the context-aware question and answer representations and their combined element-wise product (`Stream Processor' in Figure \ref{model}, e.g. for subtitles \textit{S}, the stream processor outputs [$F^{sub}$;$A^{sub,q}$;$A^{sub,a_{0-4}}$;
$F^{sub}\odot A^{sub,q}$;$F^{sub}\odot A^{sub,a_{0-4}}$]$\in\mathbb{R}^{n_{sub}\times1500}$ where $F^{sub}\in\mathbb{R}^{n_{s}\times300}$. Each concatenated vector is processed by their own unique bi-directional LSTM of output dimension 600, followed by a pair of fully connected layers of output dimensions 500 and 5, both with dropout 0.5 and ReLU activation. The 5-dimensional output represents a vote for each answer. The element-wise sum of each activated feature stream is passed to a softmax producing the predicted answer ID. All features remain separate through the entire network, effectively allowing the model to choose the most useful features. This makes this model a strong tool in assessing the biases towards certain features in the dataset.

\section{Results and Discussion}
\begin{table}[h]
\caption{(a): Each experiment is a separate end-to-end model. E.g. `SI with BERT' is the submodel of subtitles and ImageNet features (green and pink in Figure \ref{model}) with BERT embeddings used for the subtitles, questions and answers (random choice accuracy is 20\%). Models shown in \textbf{bold} surpass the SOTA. Models use timestamp annotations except \textit{STAGE} which instead uses `temporal supervision' (b): The percentages of questions in the validation set that are correctly answered by models in Group A, but incorrectly answered by Group B. \textit{Subtitle}= all models trained with subtitles. S = subtitle-only model. SVIR = model trained with all 4 features. S, V, I, R = group of 4 models each trained with one of the 4 features.}
    \begin{minipage}[t]{.5\textwidth}
        \vspace{0pt}
        \centering
            \scalebox{0.76}{
        \begin{tabular}{l|c|c|c} 
          \textbf{Model} & \textbf{Text} & \textbf{Val Set} & \textbf{Train Set}\\
          \hline\hline
          V & GloVe & 45.39\% & 60.82\%\\
          V & BERT  & 43.44\% & 52.76\%\\\hline
          I & GloVe & 44.86\% & 61.52\%\\
          I & BERT  & 44.44\% & 65.02\%\\\hline
          R & GloVe & 42.36\% & 54.83\%\\
          R & BERT  & 42.53\% & 53.85\%\\\hline
          VIR & GloVe & 46.72\% & 61.10\%\\
          VIR & BERT  & 44.61\% & 61.38\%\\\hline\hline
          S & GloVe & 66.07\% & 76.42\%\\
          S & BERT  & 68.30\% & 80.77\%\\\hline
          SI & GloVe & 67.78\% & 78.78\%\\
          \textbf{SI} & \textbf{BERT}  & \textbf{70.56\%} & 84.84\%\\\hline
          SVI & GloVe & 69.34\% & 78.90\%\\
          \textbf{SVI} & \textbf{BERT}  & \textbf{72.13\%} & 86.84\%\\\hline
          SVIR & GloVe & 69.53\% & 80.08\%\\
          \textbf{SVIR} & \textbf{BERT}  & \textbf{71.80\%} & 81.58\%\\\hline\hline
          \textit{STAGE} \cite{lei2019tvqa} & GloVe & 66.92\% & - \\
          \textit{STAGE} \cite{lei2019tvqa} & BERT & 70.50\% & - \\\hline\hline
          \textit{VSQA} \cite{Yang2020BERTRF} & GloVe & 67.70\% & -\\
          \textbf{\textit{VSQA}} \cite{Yang2020BERTRF} & \textbf{BERT} & \textbf{72.41\%} & -\\\hline\hline
          \textit{Human}  & - & 93.44\% & - \\ 
        \end{tabular}}
        \\(a)\label{tab:joint_main:main}
    \end{minipage}
    \begin{minipage}[t]{.5\textwidth}
        \vspace{0pt}
        \centering
        \scalebox{0.76}{
        \begin{tabular}{c|c|c|c} 
          \textbf{Group A} & \textbf{Group B} & \textbf{BERT} & \textbf{GloVe}\\\hline\hline
          \textit{All}          & -                              & 90.12\% & 87.68\% \\
          \textit{All}          & \textit{Non-Subtitle}          & 22.68\% & 22.91\% \\
          \textit{All}          & SVIR                           & 18.32\% & 18.16\% \\
          \textit{All}          & S, V, I, R                     &  7.68\% &  5.40\% \\\hline
          \textit{Subtitle}     & -                              & 84.74\% & 80.56\% \\
          \textit{Subtitle}     & \textit{Non-Subtitle}          & 22.68\% & 22.91\% \\\hline
          \textit{Non-Subtitle} & -                              & 67.44\% & 64.77\% \\
          \textit{Non-Subtitle} & \textit{Subtitle}              &  5.38\% &  7.12\% \\
          \textit{Non-Subtitle} & S                              & 16.72\% & 18.50\% \\\hline
          S, V, I, R                   & -                              & 82.44\% & 82.28\% \\
          S, V, I, R                   & SVIR                           & 14.44\% & 14.77\% \\\hline
          SVIR                         & -                              & 71.80\% & 69.52\% \\
          SVIR                         & S, V, I, R                     &  3.79\% &  2.01\% \\\hline
          S                            & -                              & 68.30\% & 66.07\% \\
          S                            & \textit{Non-Subtitle}          & 17.59\% & 19.80\% \\
          S                            & V, I, R                        & 21.94\% & 22.27\% \\
          S                            & VIR                            & 32.83\% & 30.78\% \\
        \end{tabular}}
        \\(b)\label{tab:joint_main:IEM}
    \end{minipage}

\label{tab:joint_main}
\end{table}

\subsection{Feature Contributions}
Table \ref{tab:joint_main}\textcolor{red}{a} shows that all evaluated models trained with subtitles significantly outperform models trained without them, with a $p-value = 1.4e-15 < 0.05$ using t-test.

\noindent \textbf{Models with Subtitles:} Each BERT variation that includes subtitles gains at least 2\% accuracy compared to GloVe, leading to the SI, SVI, and SVIR variations achieving state-of-the-art results. Models trained using only subtitles achieve +20\% accuracy using GloVe and +23\% using BERT embeddings when compared to the best performance using any combination of video features. This implies that the subtitles are the most informative features in answering the majority of the questions.

\noindent \textbf{Models without Subtitles:} With GloVe embeddings, the most impactful video feature is the visual concepts, which increases performance by 0.5\%, following a trend in image captioning \cite{inproceedings223}. Similarly, we find that using image features and visual concepts combined outperforms using either alone. However, using BERT with just visual concepts drops performance by $\sim$2\%. We theorise this is due to the contextual nature of the BERT embeddings hindering the model by sequentially processing the intrinsically unordered visual concepts.

\subsection{Subtitles Dominate Instead of Complement}
\begin{figure}[h]
    \centering
    \includegraphics[scale=0.10]{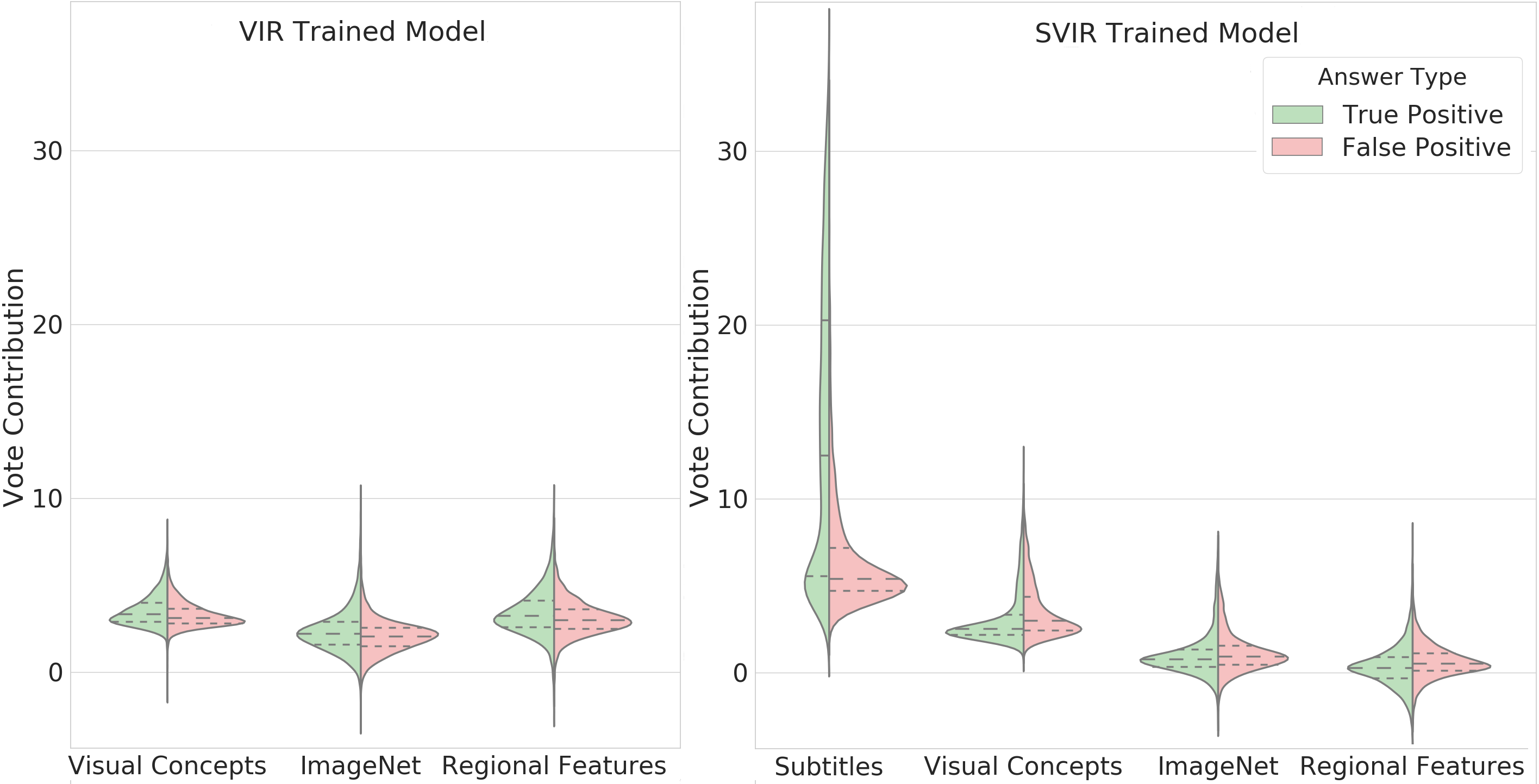}
    \caption{Pre-softmax vote contributions for answers in the validation set. Both are BERT models: VIR (left) vs SVIR (right). The dashed lines represent quartiles.}
    \label{tpfp}
\end{figure}
\begin{figure}
\centering
\begin{minipage}{.5\textwidth}
      \centering
      \includegraphics[width=0.92\linewidth]{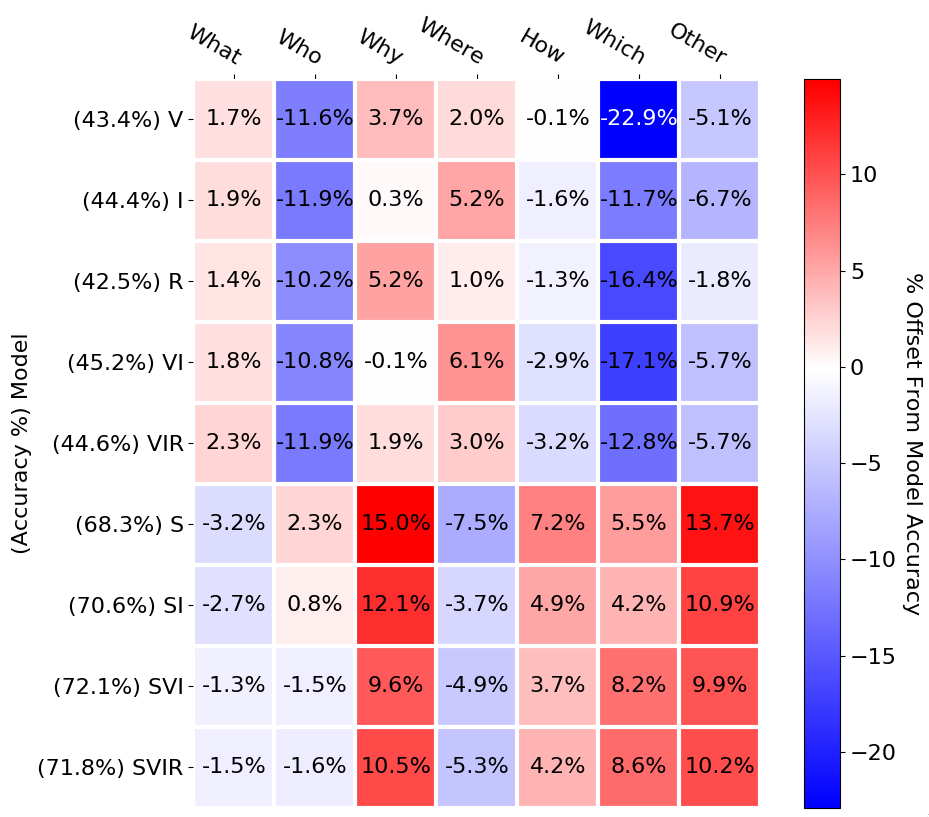}
      \\(a)\label{fig:joint_qtype_iou:qtype}
    \end{minipage}%
    \begin{minipage}{.50\textwidth}
      \centering
      \includegraphics[width=\linewidth]{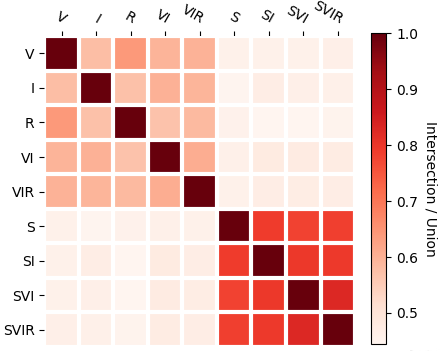}
      \\(b)\label{fig:joint_qtype_iou:iou}
    \end{minipage}
    \caption{(a) Performance of models on each question type (offset from each model's overall accuracy). (b) IoU of correct answers between models.}
    \label{fig:joint_qtype_iou}
\end{figure}

\noindent To further analyse the contributions of each feature, we plot the pre-softmax votes for answers between models trained with and without subtitles. Figure \ref{tpfp} shows the votes per feature for SVIR and VIR trained model with BERT embeddings, measured in true and false positive answers. In the VIR model (left side of Figure \ref{tpfp}) we find that all video features similarly contribute to answer votes. When averaged across correct predictions, each feature contributes positively to the correct answer i.e. true positives, and contributes less to the other incorrect answers. However, when trained with subtitles in the SVIR model, the subtitles overwhelmingly contribute to the correct answer. Furthermore, in the true positive case, each video feature actually contributes \textit{less} on average than in the accompanying incorrect answers. This is shown in the SVIR model in Figure \ref{tpfp}, where the quartiles of true positive votes in each video feature are \textit{below} the votes for false positives. This shows that in case of correct predictions, models trained with subtitles learn, on-average, to suppress the video feature contribution, demonstrating a significant bias towards subtitles in TVQA. We find similar results in the GloVe models (see supplementary materials). Strictly speaking, video-QA models that can constructively use video information at all are, to an extent, multimodal as they interpret the video features with respect to the textual questions and answers. However in TVQA, using subtitles `on-average' actively suppresses these multimodal contributions.

\subsection{All You Need is BERT}
The state-of-the-art STAGE model \cite{lei2019tvqa}, proposed by the authors of TVQA, improves on the original TVQA model by exploiting spatio-temporal relationships and simultaneously replacing GloVe with BERT embeddings. This comes at a significant increase of model complexity with over 14 additional layers and steps added to the original model. We show in Table \ref{tab:joint_main:main}\textcolor{red}{a} that simply upgrading GloVe to BERT embeddings in the relatively simple original model outperforms the more complex STAGE model. \citet{Yang2020BERTRF} present a detailed analysis of BERT on the TVQA dataset and propose a `V+S+Q+A' model that is structurally similar to the simple TVQA baseline (i.e. separate visual and subtitle streams that additively combines their contributions at the voting stage). Though they do not explore bias in TVQA, they too demonstrate a substantial boost in performance over STAGE by upgrading from GloVe to BERT on a simpler model. This implies that better modelling of the subtitles using BERT in TVQA leads to higher performance regardless of any improvement in modelling the video information. Furthermore, these results indicate that complex models focused on improving more abstract behaviours do not necessarily improve video-QA performance in TVQA. We theorise that these complex models are currently introducing unhelpful overhead and that, if the goal is to increase performance on TVQA, models are best served by exploiting the subtitles. These results also suggest that there is an imbalance in the information contributed by visual and textual modalities. We argue that the contextual nature of BERT embeddings makes them ideal for processing the sequential subtitles which, since TVQA is based on TV shows, often follow a structured narrative. 

\begin{figure}
    \centering
    \includegraphics[scale=0.14]{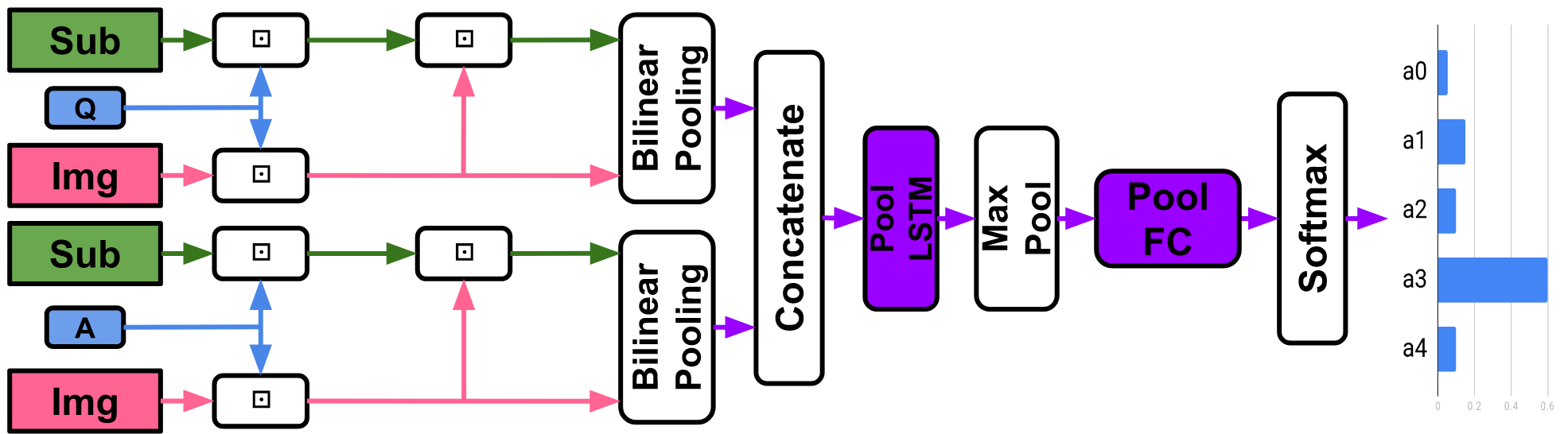}
    
    \caption{The dual-stream model. Both features are integrated into a single adapted `stream processor'. $\boxdot$ = context matching. BLP is used to fuse S and I features.\label{dual_stream_model}}
\end{figure}
\begin{table}
\caption{(a): Dual-stream vs TVQA SI baseline. The hidden pooling dimension is 1500. (b): TVQA SI model trained on the RUBi criterion (provided by \cite{Cadne2019RUBiRU}).}
    \begin{minipage}[t]{.36\textwidth}
        \vspace{0pt}
        \centering
            \scalebox{0.76}{
            \begin{tabular}{l|c|c}
              \textbf{Model} & \textbf{Text} & \textbf{Val Acc}\\
              \hline\hline
                TVQA SI & GloVe & 67.78\%\\
                TVQA SI & BERT & 70.56\%\\\hline
                Dual-Stream MCB & GloVe & 63.46\%\\
                Dual-Stream MCB & BERT & 60.63\%\\\hline
                Dual-Stream MFH & GloVe & 62.71\%\\
                Dual-Stream MFH & BERT & 59.34\%\\
            \end{tabular}}
            \\(a)\label{fig:sub:subfigure1}
    \end{minipage}
    \begin{minipage}[t]{.64\textwidth}
        \vspace{0pt}
        \centering
        \scalebox{0.76}{
        \begin{tabular}{l|c|c|c} 
          \textbf{Model} & \textbf{Dataset} & \textbf{Baseline Acc} & \textbf{RUBi Acc}\\
          \hline\hline
            TVQA SI (GloVe) & TVQA & \textbf{67.78\%} & 67.67\%\\
            TVQA SI (BERT) & TVQA & \textbf{70.56\%} & 70.37\%\\\hline
            RUBi Baseline \cite{Cadne2019RUBiRU} & VQA-CP v2 test & 38.46\% & \textbf{47.11\%}\\
            SAN \cite{Yang2015StackedAN} & VQA-CP v2 test & 33.29\% & \textbf{36.69\%}\\
            UpDn \cite{Anderson2018BottomUpAT} & VQA-CP v2 test & 41.17\% & \textbf{44.23\%}\\\hline
            RUBi Baseline \cite{Cadne2019RUBiRU} & VQA v2 test-dev & \textbf{63.10\% }& 61.16\%\\\hline
            RUBi Baseline \cite{Cadne2019RUBiRU} & VQA v2 val & \textbf{64.75\%} & 63.18\%\\
            \end{tabular}}
            \\(b)\label{fig:sub:subfigure2}
    \end{minipage}

\label{tab:joint:ds_rubi}
\end{table}

\subsection{Dataset Analysis}
\textbf{Feature Distributions:} We analyse the features that are most useful in answering each of the question types. Figure \ref{fig:joint_qtype_iou}\textcolor{red}{a} shows that models trained without subtitles significantly underperform (relative to their own model accuracy) on `which' and `who' questions. This makes intuitive sense as names and named entities commonly appear in the subtitles. Subtitle models significantly overperform on `why' and `how' questions, at $\sim$82\%. Intuitively these question types are harder because the answers are implied rather than concrete and often revolve around explanations that are best represented in language.


\begin{figure}
    \centering
    \includegraphics[width=0.55\columnwidth]{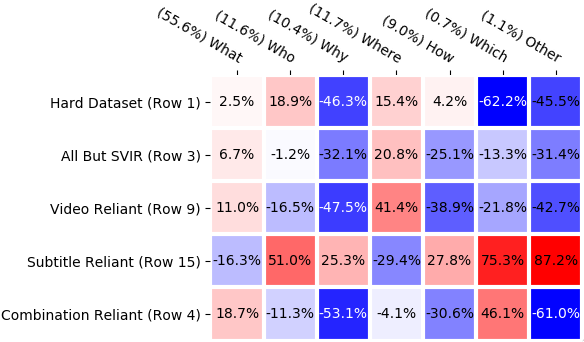}
    \caption{The percentage increase of each respective question type, in the specified IEM subset, compared to the overall question type distribution in the validation set. Each of the subsets analysed corresponds to a row in Table \ref{tab:joint_main:main}\textcolor{red}{b}.}
    \label{qtype_sets}
\end{figure}
\noindent\textbf{Modality Subsets:} By analysing the similarity between the outputs of the different models, we label each question with the modalities needed to answer it. We isolate subsets of the validation set that are answered correctly using each of the evaluated models. Figure \ref{fig:joint_qtype_iou}\textcolor{red}{b} shows that the correctly answered questions of models trained without subtitles have a relatively low intersection over union (IoU) score, approximately 58-68\%. Although the models have similar overall accuracies, they seem to perform well on different questions, implying they successfully use information from relatively separate feature types. The overall accuracies of subtitle models are significantly higher, giving higher IoU scores among those models. To inform our recommendation on how to introduce the data subsets, we run a comparative analysis of the outcomes of different groups of models, i.e. Group A and Group B, to identify the proportion of the dataset that is answered correctly by models in Group A and incorrectly by all models in Group B. We will refer to this measure as the Inclusion-Exclusion Measure (IEM). Table \ref{tab:joint_main}\textcolor{red}{b} summarises this analysis. Row `\textit{All/-}' shows that the union of all correct answers of all models is 90.1\%, whereas `\textit{All/Non-Subtitle}', contrasts predictions of all models trained with subtitles versus those trained without subtitles. This illustrates that 22.7\% of the questions cannot be answered by non-subtitle models. `\textit{Non-Subtitle/Subtitle}' shows that only 5.4\% can be uniquely answered by the models that don't use subtitles. To identify multimodal reliant questions, we consider those that SVIR can answer correctly but that the unimodal models cannot. `SVIR/S,V,I,R' shows that 3.79\% of the validation set and 2.62\% of the training set (see supplementary materials) fits this multimodal criteria ($\sim$4.3k). IEM is a strict and minimal lower bounding method. A less strict method of partitioning the dataset is to consider `popular vote', i.e. a set where the majority vote of the models in question agree on the answer. Though more restrictive than popular vote, IEM removes potential ambiguity, i.e. if a question cannot be correctly answered by any subtitle model, then subtitle content is not answering that question. Note that our proposed subsets are inherently linked to our model. Using our IEM approach, we discount the large amount of questions answered by unimodal models as not multimodal (by definition), providing a valuable starting point. Including better models as they are developed in IEM would provide increasingly better subset splits. To provide insight into how TVQA question information is actually distributed, we present the relative abundance of each question type in our proposed IEM subsets in Figure \ref{qtype_sets}. Most notably, `who' and `which' questions are more highly concentrated in the `subtitle reliant' subset. This is unsurprising as the subtitles contain a wealth of named entities and nouns. Conversely, the `video reliant' dataset contains more `what' and `where' questions.

\subsection{Further Experimental Findings}
\noindent \textbf{Joint Representations Appear Detrimental:} \citet{Baltruaitis2019MultimodalML} consider representation as summarising multimodal data ``in a way that exploits the complementarity and redundancy of multiple modalities''. Joint representations (e.g. concatenation, bilinear pooling \cite{DBLP:journals/corr/GaoBZD15, DBLP:journals/corr/FukuiPYRDR16, DBLP:journals/corr/KongF16, DBLP:journals/corr/KimOKHZ16, ben2019block}) combine unimodal signals into the same representation space. However, they struggle to handle missing data \cite{Baltruaitis2019MultimodalML} as they tend to preserve shared semantics while ignoring modality-specific information \cite{8715409}. We explore how a joint representation in the TVQA model affects performance as another method of inferring potential unimodal reliances. We create our `dual-stream' (Figure \ref{dual_stream_model}) model from the SI TVQA baseline model with as few changes as possible, i.e. I) we use context matching between subtitle and ImageNet features to allow bilinear pooling at each time step between both modalities, II) we use the new pooled feature as input for a single stream processor. Table \ref{tab:joint:ds_rubi}\textcolor{red}{a} shows that both our dual-stream models perform significantly worse than the baseline model. This implies that questions in TVQA do not effectively use a joint representation of its features, and potentially highlights: I) information from either modality is consistently missing, II) prioritising `shared semantics' over `modality-specific' information harms performance on TVQA. Both of these possibilities would contradict TVQA's stated aim as a multimodal benchmark.

\noindent \textbf{RUBi Doesn't Help:} Strong unimodal language biases are prevelent in VQA. As discussed earlier, the VQA-CP v1/2 dataset \cite{Agrawal2018DontJA} is a rearrangement of VQA v1/2 datasets \cite{Agrawal2015VQAVQ, goyal2017making} respectively such that certain kinds of identified QA priors appear exclusively in the training or test sets. Unable to rely on these priors, many VQA baseline model's performance significantly drops. The model-agnostic RUBi strategy \cite{Cadne2019RUBiRU} uses a text-only variant of a model during training (see supplementary materials for an illustration) to reduce (increase) the loss, and therefore importance, of highly-biased (visually dependent and difficult) training samples. Shown in Table \ref{tab:joint:ds_rubi}\textcolor{red}{b} , benchmark models using RUBi perform significantly better on the less-textually-biased VQA-CP dataset, implying RUBi successfully discourages models from relying on the now unhelpful text prior shortcuts. Conversely, RUBi harms performance on datasets with greater text biases (VQA v2 test-dev/val), implying RUBi's bias-averse behaviour is actually detrimental on datasets where these shortcuts exist. We find that RUBi fails to improve accuracy on TVQA and in fact \textit{slightly} decreases performance on both BERT and GloVe models, implying that TVQA could benefit heavily as a multimodal benchmark by addressing its own textual priors. To the best of our knowledge, we are the first to apply RUBi to a video-QA dataset. We note that subtitles can also provide learned text-prior shortcuts, as such, the TVQA text-only model in the RUBi strategy also includes subtitles.

\section{Conclusion}
We develop a multimodal evaluation framework using the TVQA model to assess potential dataset biases. We find that information needed to answer questions in the TVQA dataset is concentrated in the subtitles to the extent that video information is suppressed during training, contradicting the multimodal nature TVQA was specially designed to have. We provide an extensive analysis on which question types in TVQA require video or textual features and propose subsets of TVQA, in particular those which require specific features for multimodal reasoning. We achieve state-of-the-art results on the TVQA dataset by simply using BERT embeddings with the TVQA model. This demonstrates that the performance increase in the STAGE model is largely due to improved NLP embeddings. We find further evidence for TVQA's unimodal textual bias through our experiments with joint representations and the RUBi learning strategy. We show that multimodality is not always guaranteed in video-QA and suggest that it is challenging to design questions without introducing biases that discourage multimodality. We suggest that great care should be taken in creating datasets for multimodal reasoning in video-QA. To further advance the field, any research based on multimodal datasets should consider evaluating the dataset and presenting the results based on subsets and discuss how a given model performs under different modality conditions.\\

\noindent\textbf{Acknowledgements} 
This work is supported by the European Regional Development Fund. We would like to thank Liz White for her support.

\section{Supplementary Materials}
\subsection{Experimental Setup}
Our experiments are on the TVQA dataset and we use and adapt the code provided by the authors\footnote{https://github.com/jayleicn/TVQA}. Due to their size, the regional features are unavailable for download and we extract them ourselves following the author's instructions. The models are trained on an RTX 2080 Ti GPU with batch size 32 and a rectified-Adam solver \cite{liu2019variance}. We use a pretrained, non-finetuned BERT embedding layer using the uncased base tokenizer\footnote{https://github.com/huggingface/transformers}. When using regional features we use the top 20 detections per video segment. All further settings are as described in TVQA, most notably: We use 6B-300d GloVe embeddings and all word embedding layers are frozen during training. We use the timestamps annotations and train the model until improvements on the validation set accuracy is not made for 3 epochs. We check validation and training set accuracies every 400 iterations, except for the models that include regional features where we check every 800 iterations as these run significantly slower. In this study we control for the modality used in order to isolate its influence on the performance of the overall model. Details of the different variations are evaluated and their associated results are discussed in the next section. 

\subsection{Model Similarities}
\begin{figure}[h]
    \centering
    \includegraphics[width=0.5\columnwidth]{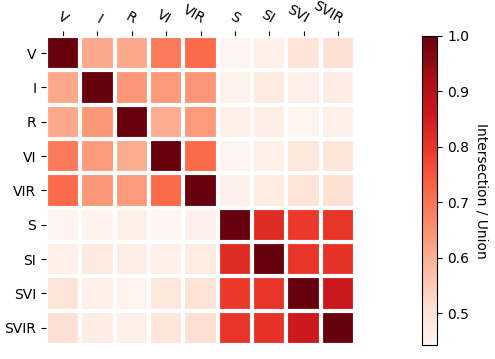}
    \caption{Intersection / Union (IoU) score for correct predictions in the validation set between GloVe models.}
    \label{glove_iou}
\end{figure}

\noindent We provide the IoU scores between the GloVe embedding model variations. The IoU scores in Figure \ref{glove_iou} are similar to their BERT counterparts shown in Figure 3b.
As an alternative set comparison measure, we consider the proportion of questions in the validation set that each pair of models answer the same, regardless if the answer is correct or incorrect.
\begin{figure}[h]
    \centering
    \includegraphics[width=0.5\columnwidth]{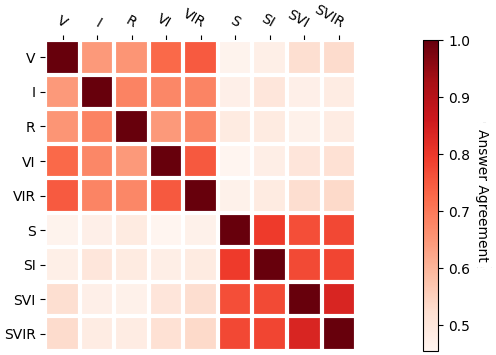}
    \caption{Proportion of the validation set that GloVe models answer the same.}
    \label{glove_agree}
\end{figure}
\begin{figure}[h]
    \centering
    \includegraphics[width=0.5\columnwidth]{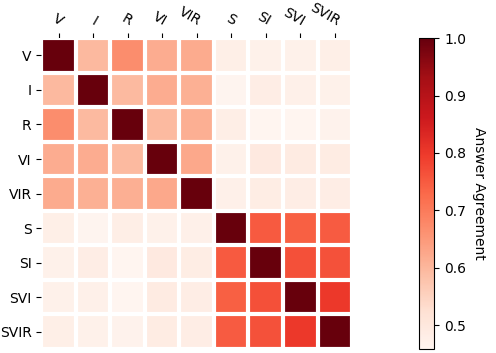}
    \caption{Proportion of the validation set that BERT models answer the same.}
    \label{bert_agree}
\end{figure}
\\We find the non-subtitle models with GloVe embeddings (Figure \ref{glove_agree}) agree \textit{slightly} more than those with BERT embeddings (Figure \ref{bert_agree}).
\onecolumn

\subsection{Question Type Analysis}
\begin{table}[h]
  \begin{center}
    \begin{tabular}{c|c} 
      \textbf{Other `Type'} & \textbf{Example} \\\hline\hline    
      Spelling Variation&`\textit{Whom} did Roger say\\& was following him after\\& he made the drop?'\\\hline
      Typo&`\textbf{tWhat} was the reason\\& House said they should\\& do a brain biopsy when\\& they were discussing\\& options of what to do?'\\\hline
      \textit{Did}/\textit{Does}&`\textit{Did} Joey walk into\\& the room before or\\& after Chandler?'\\\hline
      Double \textit{`When'}\\Question &`\textit{When} did Lucas say\\& he made the video\\& \textit{when} he was showing\\& to Beckett and Castle?'\\
    \end{tabular}
    \caption{Example questions from `other' question type category. The `other' category makes up 1.1\% of the validation set.}
    \label{other_examples}
  \end{center}
\end{table}


\subsection{Feature Contributions}
To complement the true positive and false positive vote contributions analysed in Figure 2, we present the answer vote contributions of true negative and false negative answers between VIR and SVIR trained models with both BERT and GloVe embeddings.
\begin{figure}[h]
    \centering
    \includegraphics[width=0.7\columnwidth]{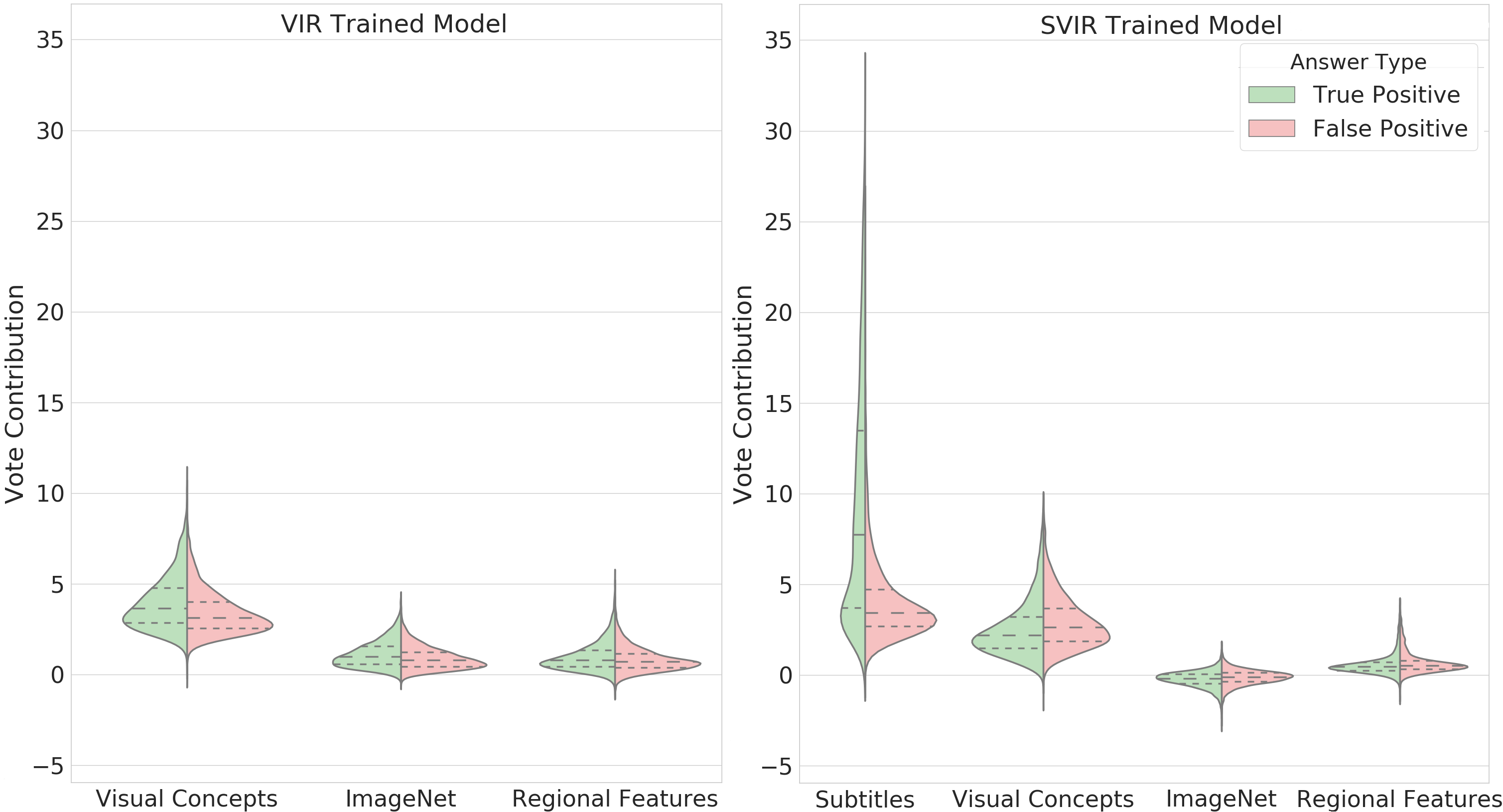}
    \caption{Pre-softmax vote contributions for answers in the validation set for the VIR (left) and SVIR (right) trained models with GloVe embeddings. This is the GloVe embedding counterpart to Figure 2.}
\end{figure}

\begin{figure}[h]
    \centering
    \includegraphics[width=0.7\columnwidth]{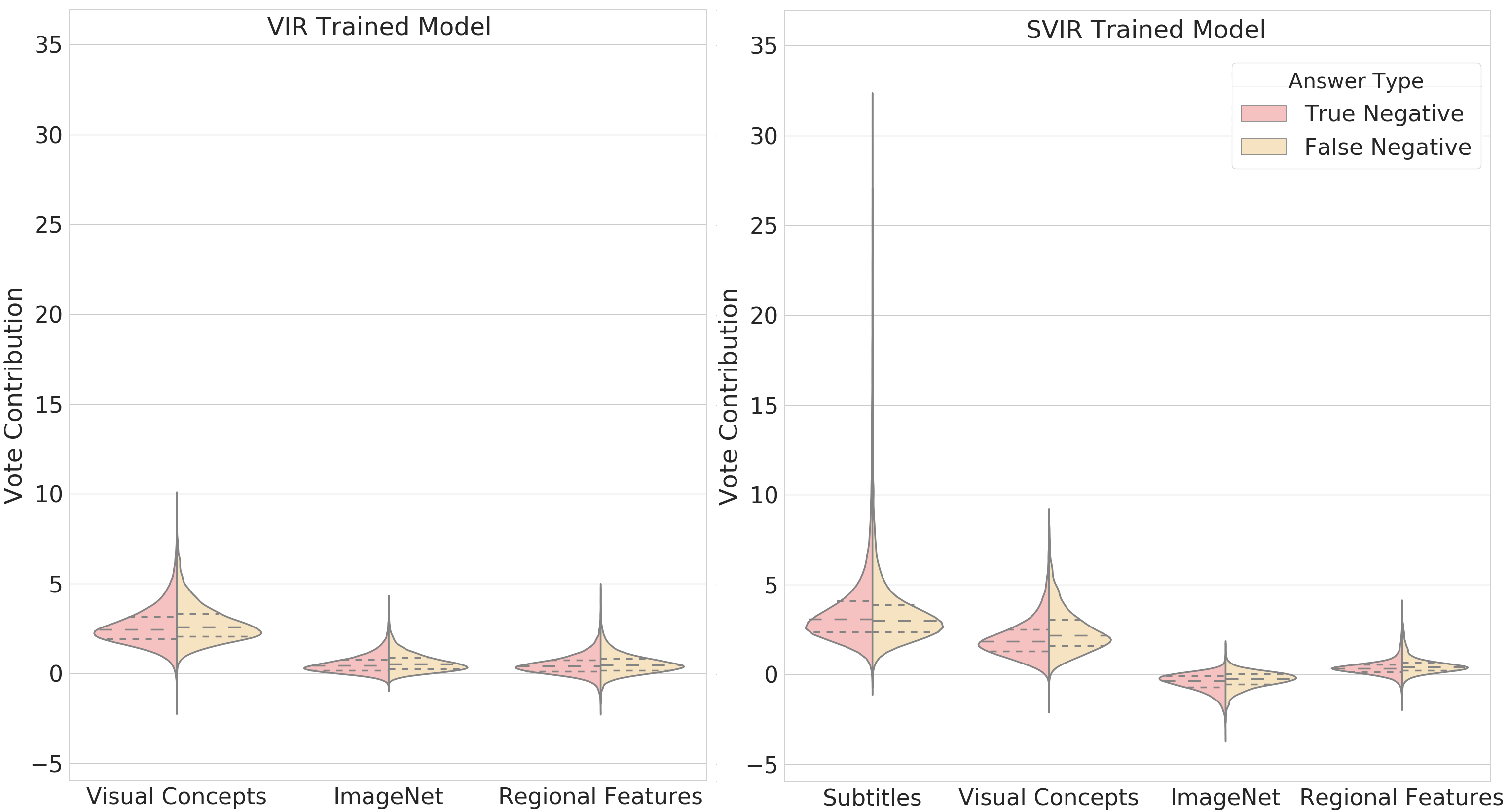}
    \caption{Pre-softmax vote contributions for answers in the validation set for the VIR (left) and SVIR (right) trained models with GloVe embeddings.}
\end{figure}

\begin{figure}[h]
    \centering
    \includegraphics[width=0.7\columnwidth]{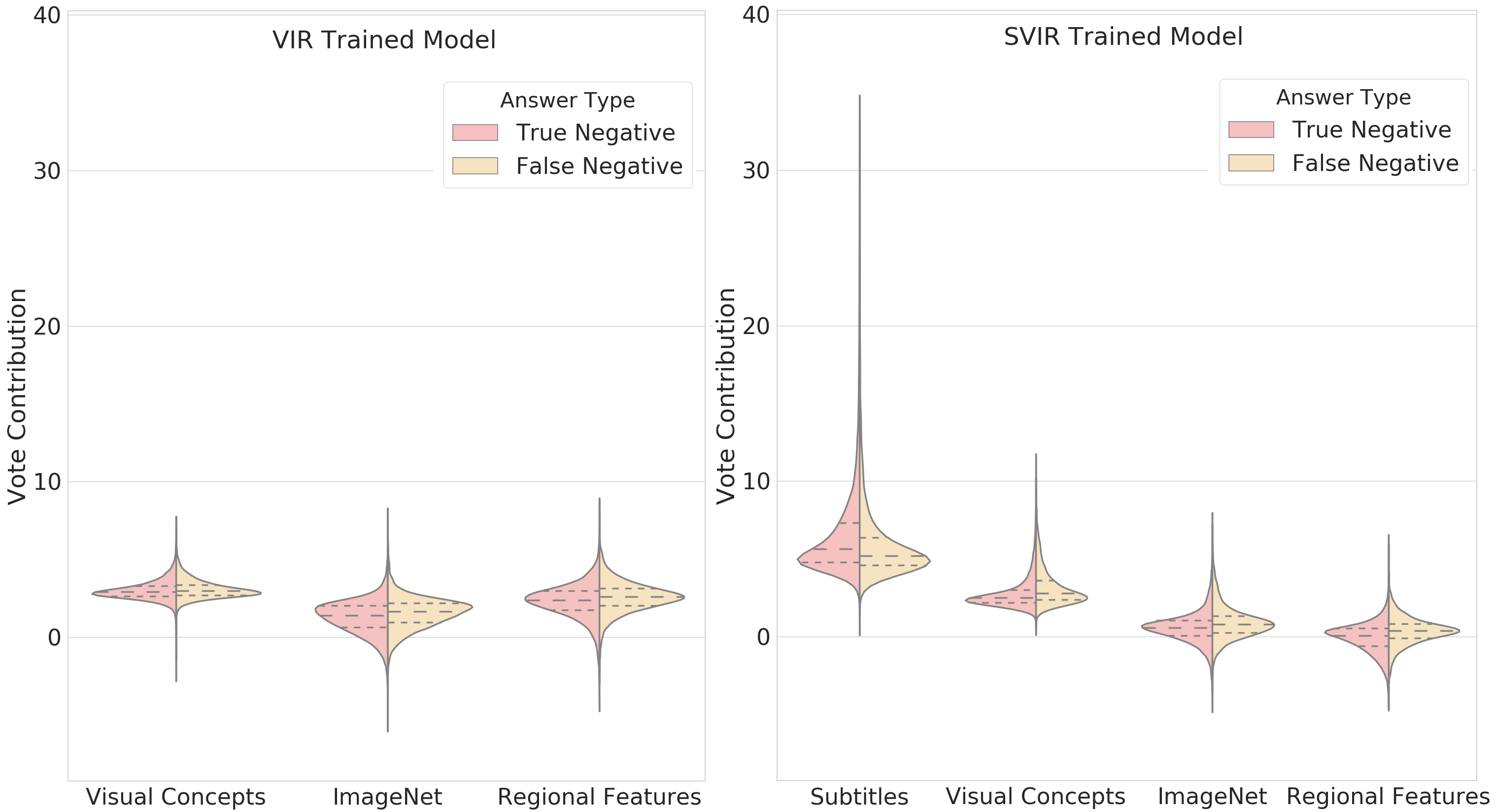}
    \caption{Pre-softmax vote contributions for answers in the validation set for the VIR (left) and SVIR (right) trained models with BERT embeddings.}
\end{figure}
\clearpage
\subsection{Training Set Inclusion-Exclusion}
\begin{table}[h]
  \begin{center}
    \begin{tabular}{c|c|c|c} 
      \textbf{Group A} & \textbf{Group B} & \textbf{BERT Models} & \textbf{GloVe Models}\\\hline\hline
      \textit{All}          & -                              & 96.77\% & 94.54\% \\\hline
      \textit{All}          & \textit{Non-Subtitle}          & 14.32\% & 14.56\% \\\hline
      \textit{All}          & SVIR                           & 15.19\% & 14.47\% \\\hline\hline
      \textit{Subtitle}     & -                              & 94.80\% & 89.91\% \\\hline
      \textit{Subtitle}     & \textit{Non-Subtitle}          & 14.32\% & 14.56\% \\\hline\hline
      \textit{Non-Subtitle} & -                              & 82.45\% & 79.99\% \\\hline
      \textit{Non-Subtitle} & \textit{Subtitle}              &  1.96\% &  4.63\% \\\hline
      \textit{Non-Subtitle} & S                              & 12.34\% & 15.97\% \\\hline\hline
      S, V, I, R                   & -                       & 91.11\% & 90.52\% \\\hline
      S, V, I, R                   & SVIR                    & 12.15\% & 12.03\% \\\hline\hline
      SVIR                         & -                       & 81.58\% & 80.07\% \\\hline
      SVIR                         & S, V, I, R              &  2.62\% &  1.58\% \\\hline\hline
      S                            & -                       & 80.77\% & 76.41\% \\\hline
      S                            & \textit{Non-Subtitle}   & 10.67\% & 12.39\% \\\hline
    \end{tabular}
    \caption{The percentages of the \textit{training} set that are correctly answered by models in Group A, but incorrectly answered by Group B. \textit{Subtitle models} = \{S, SI, SVI, SVIR\}, \textit{Non-Subtitle models} = \{V, I, R, VI, VIR\}. \textit{All models} = \textit{Subtitle} + \textit{Non-Subtitle}. Though considering responses of the training set is inherently flawed due to training bias, it provides a reasonable starting point and considerable size boost to  our initially proposed IEM subsets.}
    \label{table_innotin}
  \end{center}
\end{table}
\clearpage
\subsection{RUBi Learning Strategy}
\begin{figure}[h]
    \centering
    \includegraphics[width=0.9\columnwidth]{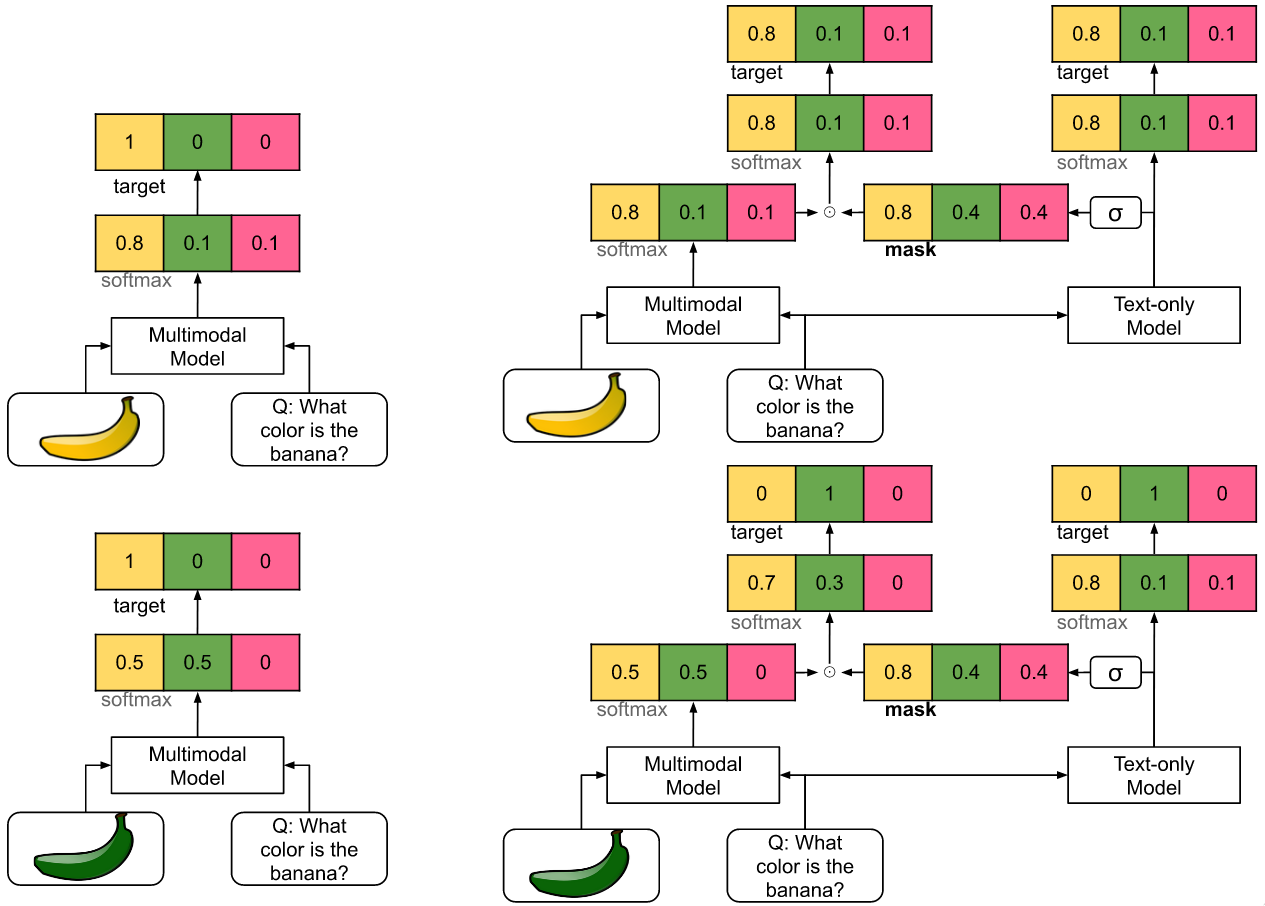}
    \caption{The RUBi (reducing unimodal bias) learning strategy used in VQA. The model-agnostic RUBi strategy \cite{Cadne2019RUBiRU} uses a text-only variant of a model during training  to reduce (increase) the loss, and therefore importance, of highly-biased (visually dependent and difficult) training samples.}
    \label{rubi_strat}
\end{figure}

\bibliography{main}

\begin{thebibliography}{51}
\providecommand{\natexlab}[1]{#1}
\providecommand{\url}[1]{\texttt{#1}}
\expandafter\ifx\csname urlstyle\endcsname\relax
  \providecommand{\doi}[1]{doi: #1}\else
  \providecommand{\doi}{doi: \begingroup \urlstyle{rm}\Url}\fi

\bibitem[Agrawal et~al.(2015)Agrawal, Lu, Antol, Mitchell, Zitnick, Parikh, and
  Batra]{DBLP:journals/corr/AntolALMBZP15}
Aishwarya Agrawal, Jiasen Lu, Stanislaw Antol, Margaret Mitchell, C.~Lawrence
  Zitnick, Devi Parikh, and Dhruv Batra.
\newblock Vqa: Visual question answering.
\newblock \emph{International Journal of Computer Vision}, 123:\penalty0 4--31,
  2015.

\bibitem[Agrawal et~al.(2018)Agrawal, Batra, Parikh, and
  Kembhavi]{Agrawal2018DontJA}
Aishwarya Agrawal, Dhruv Batra, Devi Parikh, and Aniruddha Kembhavi.
\newblock Don't just assume; look and answer: Overcoming priors for visual
  question answering.
\newblock \emph{2018 IEEE/CVF Conference on Computer Vision and Pattern
  Recognition}, pages 4971--4980, 2018.

\bibitem[AlAmri et~al.(2019)AlAmri, Cartillier, Das, Wang, Lee, Anderson, Essa,
  Parikh, Batra, Cherian, Marks, and Hori]{AlAmri2019AudioVS}
Huda AlAmri, Vincent Cartillier, Abhishek Das, Jue Wang, Stefan Lee, Pip
  Anderson, Irfan Essa, Devi Parikh, Dhruv Batra, Anoop Cherian, Tim~K. Marks,
  and Chiori Hori.
\newblock Audio visual scene-aware dialog.
\newblock \emph{2019 IEEE/CVF Conference on Computer Vision and Pattern
  Recognition (CVPR)}, pages 7550--7559, 2019.

\bibitem[Anderson et~al.(2018)Anderson, He, Buehler, Teney, Johnson, Gould, and
  Zhang]{Anderson2018BottomUpAT}
Peter Anderson, Xiaodong He, Chris Buehler, Damien Teney, Mark Johnson, Stephen
  Gould, and Lei Zhang.
\newblock Bottom-up and top-down attention for image captioning and visual
  question answering.
\newblock \emph{2018 IEEE/CVF Conference on Computer Vision and Pattern
  Recognition}, pages 6077--6086, 2018.

\bibitem[Andreas et~al.(2015)Andreas, Rohrbach, Darrell, and
  Klein]{Andreas2015DeepCQ}
Jacob Andreas, Marcus Rohrbach, Trevor Darrell, and Dan Klein.
\newblock Deep compositional question answering with neural module networks.
\newblock \emph{ArXiv}, abs/1511.02799, 2015.

\bibitem[Antol et~al.(2015)Antol, Agrawal, Lu, Mitchell, Batra, Zitnick, and
  Parikh]{Agrawal2015VQAVQ}
Stanislaw Antol, Aishwarya Agrawal, Jiasen Lu, Margaret Mitchell, Dhruv Batra,
  C.~Lawrence Zitnick, and Devi Parikh.
\newblock {VQA}: {V}isual {Q}uestion {A}nswering.
\newblock In \emph{International Conference on Computer Vision (ICCV)}, 2015.

\bibitem[Baltrusaitis et~al.(2019)Baltrusaitis, Ahuja, and
  Morency]{Baltruaitis2019MultimodalML}
Tadas Baltrusaitis, Chaitanya Ahuja, and Louis-Philippe Morency.
\newblock Multimodal machine learning: A survey and taxonomy.
\newblock \emph{IEEE Transactions on Pattern Analysis and Machine
  Intelligence}, 41:\penalty0 423--443, 2019.

\bibitem[Ben-Younes et~al.(2019)Ben-Younes, Cadene, Thome, and
  Cord]{ben2019block}
Hedi Ben-Younes, Remi Cadene, Nicolas Thome, and Matthieu Cord.
\newblock Block: Bilinear superdiagonal fusion for visual question answering
  and visual relationship detection.
\newblock In \emph{Proceedings of the AAAI Conference on Artificial
  Intelligence}, volume~33, pages 8102--8109, 2019.

\bibitem[Cad{\`e}ne et~al.(2019)Cad{\`e}ne, Dancette, Ben-younes, Cord, and
  Parikh]{Cadne2019RUBiRU}
R{\'e}mi Cad{\`e}ne, Corentin Dancette, Hedi Ben-younes, Matthieu Cord, and
  Devi Parikh.
\newblock Rubi: Reducing unimodal biases in visual question answering.
\newblock In \emph{NeurIPS}, 2019.

\bibitem[Chao et~al.(2018)Chao, Hu, and Sha]{Chao2018BeingNB}
Wei-Lun Chao, Hexiang Hu, and Fei Sha.
\newblock Being negative but constructively: Lessons learnt from creating
  better visual question answering datasets.
\newblock In \emph{NAACL-HLT}, 2018.

\bibitem[Chen et~al.(2020)Chen, Yan, Xiao, Zhang, Pu, and
  Zhuang]{Chen2020CounterfactualSS}
Long Chen, Xin Yan, Jun Xiao, Hanwang Zhang, Shiliang Pu, and Yueting Zhuang.
\newblock Counterfactual samples synthesizing for robust visual question
  answering.
\newblock In \emph{CVPR}, 2020.

\bibitem[Das et~al.(2019)Das, Kottur, Gupta, Singh, Yadav, Lee, Moura, Parikh,
  and Batra]{Das2019VisualD}
Abhishek Das, Satwik Kottur, Khushi Gupta, Avi Singh, Deshraj Yadav, Stefan
  Lee, Jos{\'e} M.~F. Moura, Devi Parikh, and Dhruv Batra.
\newblock Visual dialog.
\newblock \emph{IEEE Transactions on Pattern Analysis and Machine
  Intelligence}, 41:\penalty0 1242--1256, 2019.

\bibitem[{Deng} et~al.(2009){Deng}, {Dong}, {Socher}, {Li}, {Kai Li}, and {Li
  Fei-Fei}]{5206848}
J.~{Deng}, W.~{Dong}, R.~{Socher}, L.~{Li}, {Kai Li}, and {Li Fei-Fei}.
\newblock Imagenet: A large-scale hierarchical image database.
\newblock In \emph{2009 IEEE Conference on Computer Vision and Pattern
  Recognition}, pages 248--255, June 2009.
\newblock \doi{10.1109/CVPR.2009.5206848}.

\bibitem[Devlin et~al.(2019)Devlin, Chang, Lee, and Toutanova]{devlin2018bert}
Jacob Devlin, Ming-Wei Chang, Kenton Lee, and Kristina Toutanova.
\newblock Bert: Pre-training of deep bidirectional transformers for language
  understanding.
\newblock In \emph{NAACL-HLT}, 2019.

\bibitem[Fan(2019)]{Fan_2019_ICCV}
Chenyou Fan.
\newblock Egovqa - an egocentric video question answering benchmark dataset.
\newblock In \emph{The IEEE International Conference on Computer Vision (ICCV)
  Workshops}, Oct 2019.

\bibitem[Fukui et~al.(2016)Fukui, Park, Yang, Rohrbach, Darrell, and
  Rohrbach]{DBLP:journals/corr/FukuiPYRDR16}
Akira Fukui, Dong~Huk Park, Daylen Yang, Anna Rohrbach, Trevor Darrell, and
  Marcus Rohrbach.
\newblock Multimodal compact bilinear pooling for visual question answering and
  visual grounding.
\newblock In \emph{Proceedings of the 2016 Conference on Empirical Methods in
  Natural Language Processing}, pages 457--468, Austin, Texas, November 2016.
  Association for Computational Linguistics.
\newblock \doi{10.18653/v1/D16-1044}.
\newblock URL \url{https://www.aclweb.org/anthology/D16-1044}.

\bibitem[Gao et~al.(2016)Gao, Beijbom, Zhang, and
  Darrell]{DBLP:journals/corr/GaoBZD15}
Yang Gao, Oscar Beijbom, Ning Zhang, and Trevor Darrell.
\newblock Compact bilinear pooling.
\newblock \emph{2016 IEEE Conference on Computer Vision and Pattern Recognition
  (CVPR)}, pages 317--326, 2016.

\bibitem[Girdhar and Ramanan(2020)]{girdhar2020cater}
Rohit Girdhar and Deva Ramanan.
\newblock {CATER: A diagnostic dataset for Compositional Actions and TEmporal
  Reasoning}.
\newblock In \emph{ICLR}, 2020.

\bibitem[Gordon and Van~Durme(2013)]{10.1145/2509558.2509563}
Jonathan Gordon and Benjamin Van~Durme.
\newblock Reporting bias and knowledge acquisition.
\newblock New York, NY, USA, 2013. Association for Computing Machinery.
\newblock ISBN 9781450324113.
\newblock \doi{10.1145/2509558.2509563}.
\newblock URL \url{https://doi.org/10.1145/2509558.2509563}.

\bibitem[Goyal et~al.(2017)Goyal, Khot, Summers-Stay, Batra, and
  Parikh]{goyal2017making}
Yash Goyal, Tejas Khot, Douglas Summers-Stay, Dhruv Batra, and Devi Parikh.
\newblock Making the v in vqa matter: Elevating the role of image understanding
  in visual question answering.
\newblock In \emph{Proceedings of the IEEE Conference on Computer Vision and
  Pattern Recognition}, pages 6904--6913, 2017.

\bibitem[Graves and Schmidhuber(2005)]{graves2005framewise}
Alex Graves and J{\"u}rgen Schmidhuber.
\newblock Framewise phoneme classification with bidirectional lstm and other
  neural network architectures.
\newblock \emph{Neural Networks}, 18\penalty0 (5-6):\penalty0 602--610, 2005.

\bibitem[{Guo} et~al.(2019){Guo}, {Wang}, and {Wang}]{8715409}
W.~{Guo}, J.~{Wang}, and S.~{Wang}.
\newblock Deep multimodal representation learning: A survey.
\newblock \emph{IEEE Access}, 7:\penalty0 63373--63394, 2019.

\bibitem[Gupta(2017)]{Gupta2017SurveyOV}
Akshay~Kumar Gupta.
\newblock Survey of visual question answering: Datasets and techniques.
\newblock \emph{CoRR}, abs/1705.03865, 2017.
\newblock URL \url{http://arxiv.org/abs/1705.03865}.

\bibitem[He et~al.(2016)He, Zhang, Ren, and Sun]{DBLP:journals/corr/HeZRS15}
Kaiming He, Xiangyu Zhang, Shaoqing Ren, and Jian Sun.
\newblock Deep residual learning for image recognition.
\newblock \emph{2016 IEEE Conference on Computer Vision and Pattern Recognition
  (CVPR)}, pages 770--778, 2016.

\bibitem[Hochreiter and Schmidhuber(1997)]{Hochreiter:1997:LSM:1246443.1246450}
Sepp Hochreiter and J\"{u}rgen Schmidhuber.
\newblock Long short-term memory.
\newblock \emph{Neural Comput.}, 9\penalty0 (8):\penalty0 1735--1780, November
  1997.
\newblock ISSN 0899-7667.
\newblock \doi{10.1162/neco.1997.9.8.1735}.
\newblock URL \url{http://dx.doi.org/10.1162/neco.1997.9.8.1735}.

\bibitem[Jang et~al.(2017)Jang, Song, Yu, Kim, and Kim]{jang-CVPR-2017}
Yunseok Jang, Yale Song, Youngjae Yu, Youngjin Kim, and Gunhee Kim.
\newblock Tgif-qa: Toward spatio-temporal reasoning in visual question
  answering.
\newblock \emph{2017 IEEE Conference on Computer Vision and Pattern Recognition
  (CVPR)}, pages 1359--1367, 2017.

\bibitem[Karpathy and Fei-Fei(2015)]{Karpathy2015DeepVA}
Andrej Karpathy and Li~Fei-Fei.
\newblock Deep visual-semantic alignments for generating image descriptions.
\newblock In \emph{CVPR}, 2015.

\bibitem[Kim et~al.(2017)Kim, On, Lim, Kim, Ha, and
  Zhang]{DBLP:journals/corr/KimOKHZ16}
Jin-Hwa Kim, Kyoung~Woon On, Woosang Lim, Jeonghee Kim, Jung-Woo Ha, and
  Byoung-Tak Zhang.
\newblock {Hadamard Product for Low-rank Bilinear Pooling}.
\newblock In \emph{The 5th International Conference on Learning
  Representations}, 2017.

\bibitem[Kim et~al.(2016)Kim, Nan, Heo, Choi, and Zhang]{pororoqa}
K~Kim, C~Nan, MO~Heo, SH~Choi, and BT~Zhang.
\newblock Pororoqa: Cartoon video series dataset for story understanding.
\newblock In \emph{Proceedings of NIPS 2016 Workshop on Large Scale Computer
  Vision System}, volume~19, 2016.

\bibitem[Kong and Fowlkes(2017)]{DBLP:journals/corr/KongF16}
Shu Kong and Charless~C. Fowlkes.
\newblock Low-rank bilinear pooling for fine-grained classification.
\newblock \emph{2017 IEEE Conference on Computer Vision and Pattern Recognition
  (CVPR)}, pages 7025--7034, 2017.

\bibitem[Krishna et~al.(2016)Krishna, Zhu, Groth, Johnson, Hata, Kravitz, Chen,
  Kalantidis, Li, Shamma, Bernstein, and Fei-Fei]{krishnavisualgenome}
Ranjay Krishna, Yuke Zhu, Oliver Groth, Justin Johnson, Kenji Hata, Joshua
  Kravitz, Stephanie Chen, Yannis Kalantidis, Li-Jia Li, David~A. Shamma,
  Michael~S. Bernstein, and Li~Fei-Fei.
\newblock Visual genome: Connecting language and vision using crowdsourced
  dense image annotations.
\newblock \emph{International Journal of Computer Vision}, 123:\penalty0
  32--73, 2016.

\bibitem[Lei et~al.(2018)Lei, Yu, Bansal, and Berg]{lei2018tvqa}
Jie Lei, Licheng Yu, Mohit Bansal, and Tamara~L Berg.
\newblock Tvqa: Localized, compositional video question answering.
\newblock In \emph{EMNLP}, 2018.

\bibitem[Lei et~al.(2020)Lei, Yu, Berg, and Bansal]{lei2019tvqa}
Jie Lei, Licheng Yu, Tamara Berg, and Mohit Bansal.
\newblock {TVQA}+: Spatio-temporal grounding for video question answering.
\newblock In \emph{Proceedings of the 58th Annual Meeting of the Association
  for Computational Linguistics}, pages 8211--8225, Online, July 2020.
  Association for Computational Linguistics.
\newblock \doi{10.18653/v1/2020.acl-main.730}.
\newblock URL \url{https://www.aclweb.org/anthology/2020.acl-main.730}.

\bibitem[Liu et~al.(2019)Liu, Jiang, He, Chen, Liu, Gao, and
  Han]{liu2019variance}
Liyuan Liu, Haoming Jiang, Pengcheng He, Weizhu Chen, Xiaodong Liu, Jianfeng
  Gao, and Jiawei Han.
\newblock On the variance of the adaptive learning rate and beyond.
\newblock \emph{arXiv preprint arXiv:1908.03265}, 2019.

\bibitem[Lu et~al.(2015)Lu, Lin, Batra, and Parikh]{Lu2015}
Jiasen Lu, Xiao Lin, Dhruv Batra, and Devi Parikh.
\newblock Deeper lstm and normalized cnn visual question answering model.
\newblock \url{https://github.com/VT-vision-lab/VQA_LSTM_CNN}, 2015.

\bibitem[Maharaj et~al.(2017)Maharaj, Ballas, Rohrbach, Courville, and
  Pal]{maharaj2017dataset}
Tegan Maharaj, Nicolas Ballas, Anna Rohrbach, Aaron~C Courville, and
  Christopher~Joseph Pal.
\newblock A dataset and exploration of models for understanding video data
  through fill-in-the-blank question-answering.
\newblock In \emph{Computer Vision and Pattern Recognition (CVPR)}, 2017.
\newblock URL
  \url{http://openaccess.thecvf.com/content_cvpr_2017/papers/Maharaj_A_Dataset_and_CVPR_2017_paper.pdf}.

\bibitem[Malinowski and Fritz(2014)]{DBLP:journals/corr/MalinowskiF14}
Mateusz Malinowski and Mario Fritz.
\newblock A multi-world approach to question answering about real-world scenes
  based on uncertain input.
\newblock In \emph{NIPS}, 2014.

\bibitem[Pennington et~al.(2014)Pennington, Socher, and
  Manning]{pennington2014glove}
Jeffrey Pennington, Richard Socher, and Christopher~D. Manning.
\newblock Glove: Global vectors for word representation.
\newblock In \emph{Empirical Methods in Natural Language Processing (EMNLP)},
  pages 1532--1543, 2014.
\newblock URL \url{http://www.aclweb.org/anthology/D14-1162}.

\bibitem[Ramakrishnan et~al.(2018)Ramakrishnan, Agrawal, and
  Lee]{Ramakrishnan2018OvercomingLP}
Sainandan Ramakrishnan, Aishwarya Agrawal, and Stefan Lee.
\newblock Overcoming language priors in visual question answering with
  adversarial regularization.
\newblock In \emph{NeurIPS}, 2018.

\bibitem[Ren et~al.(2015)Ren, He, Girshick, and
  Sun]{DBLP:journals/corr/RenHG015}
Shaoqing Ren, Kaiming He, Ross~B. Girshick, and Jian Sun.
\newblock Faster r-cnn: Towards real-time object detection with region proposal
  networks.
\newblock \emph{IEEE Transactions on Pattern Analysis and Machine
  Intelligence}, 39:\penalty0 1137--1149, 2015.

\bibitem[Seo et~al.(2017)Seo, Kembhavi, Farhadi, and
  Hajishirzi]{DBLP:journals/corr/SeoKFH16}
Min~Joon Seo, Aniruddha Kembhavi, Ali Farhadi, and Hannaneh Hajishirzi.
\newblock Bidirectional attention flow for machine comprehension.
\newblock \emph{ICLR}, abs/1611.01603, 2017.
\newblock URL \url{http://arxiv.org/abs/1611.01603}.

\bibitem[Tapaswi et~al.(2016)Tapaswi, Zhu, Stiefelhagen, Torralba, Urtasun, and
  Fidler]{MovieQA}
Makarand Tapaswi, Yukun Zhu, Rainer Stiefelhagen, Antonio Torralba, Raquel
  Urtasun, and Sanja Fidler.
\newblock {MovieQA: Understanding Stories in Movies through
  Question-Answering}.
\newblock In \emph{IEEE Conference on Computer Vision and Pattern Recognition
  (CVPR)}, 2016.

\bibitem[Tommasi et~al.(2017)Tommasi, Patricia, Caputo, and
  Tuytelaars]{tommasi2017deeper}
Tatiana Tommasi, Novi Patricia, Barbara Caputo, and Tinne Tuytelaars.
\newblock A deeper look at dataset bias.
\newblock In \emph{Domain adaptation in computer vision applications}, pages
  37--55. Springer, 2017.

\bibitem[{Torralba} and {Efros}(2011)]{5995347}
A.~{Torralba} and A.~A. {Efros}.
\newblock Unbiased look at dataset bias.
\newblock In \emph{CVPR 2011}, pages 1521--1528, 2011.

\bibitem[Yang et~al.(2020)Yang, Garcia, Chu, Otani, Nakashima, and
  Takemura]{Yang2020BERTRF}
Zekun Yang, Noa Garcia, Chenhui Chu, Mayu Otani, Yuta Nakashima, and Haruo
  Takemura.
\newblock Bert representations for video question answering.
\newblock \emph{2020 IEEE Winter Conference on Applications of Computer Vision
  (WACV)}, pages 1545--1554, 2020.

\bibitem[Yang et~al.(2015)Yang, He, Gao, Deng, and Smola]{Yang2015StackedAN}
Zichao Yang, Xiaodong He, Jianfeng Gao, Li~Deng, and Alexander~J. Smola.
\newblock Stacked attention networks for image question answering.
\newblock \emph{2016 IEEE Conference on Computer Vision and Pattern Recognition
  (CVPR)}, pages 21--29, 2015.

\bibitem[Ye et~al.(2017)Ye, Zhao, Li, Chen, Xiao, and
  Zhuang]{DBLP:journals/corr/YeZLCXZ17}
Yunan Ye, Zhou Zhao, Yimeng Li, Long Chen, Jun Xiao, and Yueting Zhuang.
\newblock Video question answering via attribute-augmented attention network
  learning.
\newblock \emph{Proceedings of the 40th International ACM SIGIR Conference on
  Research and Development in Information Retrieval}, 2017.

\bibitem[Yi* et~al.(2020)Yi*, Gan*, Li, Kohli, Wu, Torralba, and
  Tenenbaum]{Yi*2020CLEVRER:}
Kexin Yi*, Chuang Gan*, Yunzhu Li, Pushmeet Kohli, Jiajun Wu, Antonio Torralba,
  and Joshua~B. Tenenbaum.
\newblock Clevrer: Collision events for video representation and reasoning.
\newblock In \emph{International Conference on Learning Representations}, 2020.
\newblock URL \url{https://openreview.net/forum?id=HkxYzANYDB}.

\bibitem[Yin and Ordonez(2017)]{inproceedings223}
Xuwang Yin and Vicente Ordonez.
\newblock Obj2text: Generating visually descriptive language from object
  layouts.
\newblock In \emph{Proceedings of the 2017 Conference on Empirical Methods in
  Natural Language Processing}, pages 177--187, Copenhagen, Denmark, September
  2017. Association for Computational Linguistics.
\newblock \doi{10.18653/v1/D17-1017}.
\newblock URL \url{https://www.aclweb.org/anthology/D17-1017}.

\bibitem[Yu et~al.(2018)Yu, Dohan, Le, Luong, Zhao, and
  Chen]{DBLP:journals/corr/abs-1804-09541}
Adams~Wei Yu, David Dohan, Quoc Le, Thang Luong, Rui Zhao, and Kai Chen.
\newblock Fast and accurate reading comprehension by combining self-attention
  and convolution.
\newblock In \emph{International Conference on Learning Representations}, 2018.
\newblock URL \url{https://openreview.net/forum?id=B14TlG-RW}.

\bibitem[Zhang et~al.(2020)Zhang, Yang, He, and Deng]{zhang2019multimodal}
Chao Zhang, Zichao Yang, Xiaodong He, and Li~Deng.
\newblock Multimodal intelligence: Representation learning, information fusion,
  and applications.
\newblock \emph{IEEE Journal of Selected Topics in Signal Processing},
  14:\penalty0 478--493, 2020.

\end{thebibliography}
\end{document}